\PassOptionsToPackage{table}{xcolor}
\documentclass{article}
\usepackage[utf8]{inputenc}
\usepackage{amsmath}
\usepackage{amsthm}
\usepackage{amsfonts}
\usepackage{amssymb}
\usepackage{enumerate}
\usepackage{tikz}
\usepackage{youngtab}
\usepackage{mathtools}
\usepackage{blkarray}
\usepackage{empheq}
\usepackage{subfig}
\usepackage{floatrow}

\usepackage{pgfplots}
\usepackage{tabularx}
\usepackage{algorithm,algorithmic}
\usepackage{hyperref}

\usepackage[square,numbers]{natbib}
\begin{document}

\title{Ensemble Mask Networks}



\title{Ensemble Mask Networks}

\author{Jonny Luntzel}
\date{}

\maketitle

\tikzset{highlight/.style={rectangle,
        fill=red!15,
        blend mode = multiply,
        rounded corners = 0.5 mm,
        inner sep=1pt,
        fit = #1}}          
\section*{Abstract}
Can an $\mathbb{R}^n\rightarrow \mathbb{R}^n$ feedforward network learn matrix-vector multiplication? This study introduces two mechanisms - flexible masking to take matrix inputs, and a unique network pruning to respect the mask's dependency structure. Networks can approximate fixed operations such as matrix-vector multiplication $\phi(A,x) \rightarrow Ax$, motivating the mechanisms introduced with applications towards litmus-testing dependencies or interaction order in graph-based models.
\section{Introduction}
Ensembling \cite{EN} networks together, encoding their dependency structures \cite{NGM} while regressing to data, pruning \cite{will} networks, and sparse masked training \cite{FISH} find interest and development. Instead of pooling multiple networks, a singular network acts as an ensemble of masked subnetworks whose input dependencies are specified by an adjacency matrix, and trained neural networks are treated as fixed operations which can take a family of linear operators rather than fixing them. Teaching standard neural networks multiplication of scalar inputs \cite{nalu,MA} is nontrivial but has been explored, and while they clearly can regress $x \rightarrow Ax$ for a fixed matrix, the general case is less often discussed. Two key features of this basic operation - arbitrary input and row-wise multiplication - highlight how  flexible masking and a tailored architecture address these demands.

\begin{figure}[H]
{\includegraphics[width=.6\textwidth]{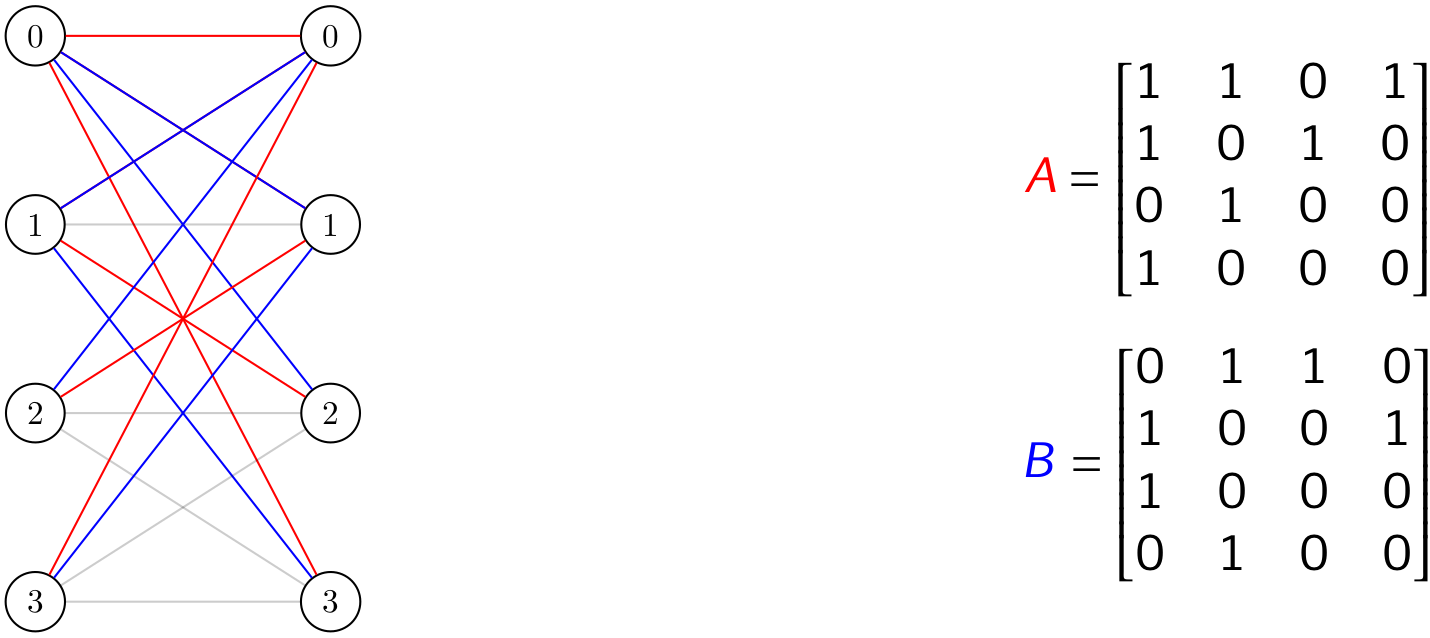}}
\caption{The network's first layer with 2 adjacency masks superimposed. It learns weights for a fully connected layer while associating all $2^n$ edge subsets with distinct matrix masks.}
\label{fig:o}
\end{figure}

\section{Network Architecture}
\begin{figure}[hbp]
\[A= \begin{bsmallmatrix}
        0 & 1 & 0 & 1\\
        0 & 0 & 1 & 0\\
        1 & 0 & 0 & 0\\
        0 & 0 & 1 & 0
    \end{bsmallmatrix}\]
  \centering
  \subfloat[(1)]{\includegraphics[width=0.24\textwidth]{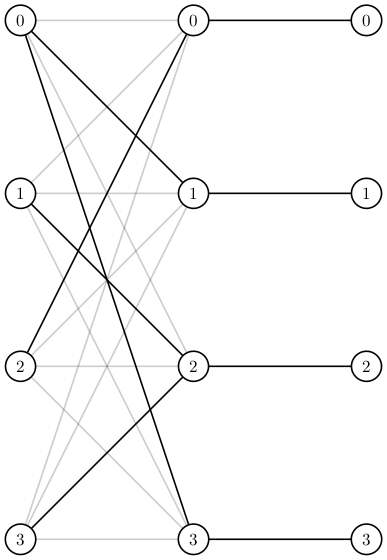}\label{fig:f1}}
  \hfill
  \subfloat[(2)]{\includegraphics[width=0.24\textwidth]{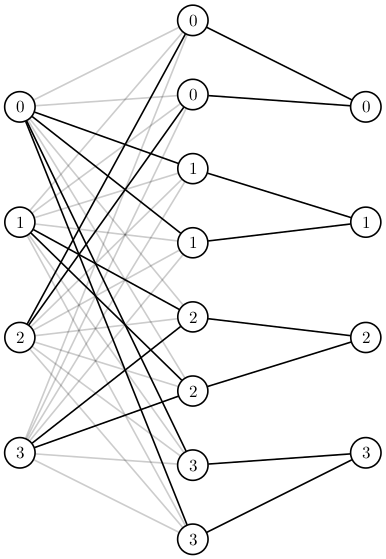}\label{fig:f2}}
  \hfill
  \subfloat[(2,3)]
  {\includegraphics[width=0.24\textwidth]{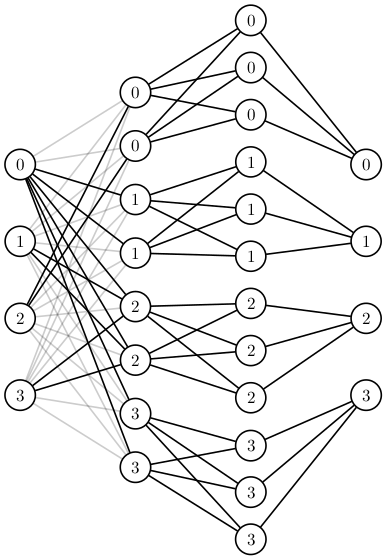}\label{fig:f3}}
\caption{Multiple network architectures. Left to right: 1 hidden layer with 1 copy per node, 1 hidden layer with 2 copies per node, and 2 hidden layers with 2 and 3 copies per node.  $A$ masks and scales edges by its non-zero values.}
\label{fig:d}
\end{figure}

An ensemble mask network (Fig. \ref{fig:d}) is a conventional feedforward neural network which restricts edges in its computation according to an adjacency matrix. By identifying $(i,j)$ matrix entries with edges connecting indexes $i$ and $j$ of the input and subsequent layer respectively, we can input $A$ and mute inactive edges. $A$ may be  directed and weighted with self-loops - unsymmetric with $A_{ij} \in [0,1]$ and diagonal values - where weights simply temporarily scale active edges in the first layer. Flexibly masking\footnote{github.com/uchida-takumi/CustomizedLinear} the first layer affords general matrix input and encourages the network to learn the target fixed operation using only necessary information up to the second layer, where further modifications prevent mixing.
\newline\\
The remaining layers are pruned to respect the target operation's input dependencies such as the first-order interactions of each row-wise multiplication, where graphically a node is a function of its direct incoming neighbors (or non-zero entries in its row of $A$). The output does not factor in empty entries to streamline the network and match the interaction order  of the target operation (sequential masks (Fig. \ref{fig:snd}) account for higher orders). Input $\&$ output layers are size $n$, and hidden layers $i$ are $c_i n$. While the first edge set is flexibly masked by $A$, subsequent edges are fixed and one-to-one in the sense that each cluster of $c_i$ nodes are associated with one of the $n$ input nodes (however they are fully connected across said clusters). This architecture respects dependencies of $A$, restricting the function space spanned by the network.
\newline
\section{Similarity to GNNs}
A neural network takes vector inputs $x \in \mathbb{R}^n$ to vector outputs $y \in \mathbb{R}^k$, while node-regressing graph neural networks broadly take vector $x \in \mathbb{R}^n$ and matrix $A \in \mathbb{R}^{nxn}$ inputs to vector outputs $y \in \mathbb{R}^n$. Ensemble mask networks fall between these two tools in terms of their expressivity because they do not explicitly take a matrix $A$ as input that can be flexibly operated on via activation functions or for nth order message passing, nor do they operate solely on $x$. They take $\mathbb{R}^n \rightarrow \mathbb{R}^n$ using $A$ to access an ensemble of $2^n$ (Fig. \ref{fig:o}) subnetworks. 
\newline\\
Graph neural networks are of the form: \[h_{u} = \phi(x_{u}, \bigoplus_{v \in N_{u}} \psi (x_{u}, x_{v}, e_{uv}))\] \newline Given inputs $x$ and hidden layer representations $h$, $\psi$ is some ``message" activation function relating node $u$ to its neighbors $v \in N_u$ by taking inputs $x_u, x_v$. After being aggregated, activation function $\phi$ updates the output $h_u$ by relating aggregated information to node input $x_u$. Because it accumulates information from immediate neighbors this determines one message passing operation.
\newline\\
For this task the aggregation operator is a row-wise multiplication (consider a node in the first hidden layer - its values are a weighted sum of the neighbors of the node it represents). Let $A_i$ be the i'th row of $A$. $W_{1}$ refers to a layer 1 weight matrix with same shape/node identities as A. $W_{1_{i}}$ is the i'th row of $W_{1}$. The first hidden layer can be written as
\[h_{u} = \phi(x_{u}, A_{u}\odot \psi(W_{1_{u}}x)).\]
We mask the weights by $A$ and each row-wise operation contributes to the MPNN layer. There is no activation function $\phi$ which relates node u to the aggregated value, nor are there intermediate activation functions $\psi$ because these experiments are proof of concept and disband activation functions, simplifying outputs to
\[h_{u} = (A_{u}\odot W_{1_{u}})x\]
To get output $y$ we compute
\[y = W_{n}\hdots W_{2}(A \odot W_{1})x\] While the first MPNN operation is $(A \odot W_{1})x$, remaining layers are identity matrices or self-loop graphs with $N_{u} = x_{u}$. Inputs and outputs are size $(d,1)$, but hidden layer dimensions may vary as the i'th hidden layer $h$ is of size $(c_{i}d, 1)$, so the feed-forward network composes matrix multiplications $(d, c_{n}d)(c_n d, (c_{n-1}d)\hdots(c_{1}d, d)(d, 1)$ where the next hidden layer $W_{i}h$ is $(c_{i+1}d,1)$.\newline\\
\newline\\

\section{Experiments}
Experiments used a hidden layer architecture $(1)$, no activation functions, training data $d$ of size $36n^{1.3}$, $k=8n$ vector samples per training datum, a batch size of $\frac{k}{4}$, SGD, and learning rate $0.1$. A parameter sweep of epochs in $[1,2,3]$ and meta-epochs in $[1,2,4,8,16]$ was conducted for sizes $\text{n} \in [5,10,20,40,100]$. Epochs denote the number of times the network trains for fixed data in a training session, while meta-epochs denote the number of sessions (Algorithm 1). For example, 2 epochs and 4 meta-epochs implies we train over $d$ graphs each with $k$ twice-repeated training vectors, composing a single training session, with 4 repeated sessions. Input vector values were uniformly sampled in $[0,1]$\footnote{The unit input vector restriction generalizes since arbitrary vectors can be normalized and outputs scaled by the input's norm.}. The training algorithm below describes how networks couple edge weights to matrix entries. An input in $d$ masks and bakes in its adjacency structure to construct a subnetwork before training. Loss computes and the first layer gradient updates unmasked edges while the remaining layers update normally. The network then merges with its masked units and replaces updated weights to prepare more vector inputs for another matrix entry. There are the experimental conditions - control fixes $d$ across meta-epochs, shuffling permutes $d$ each meta-epoch, and randomized subsampling generates a new $d$ each time.
\begin{algorithm}[H]
\begin{algorithmic}
\FOR{$i=1$ to \text{meta-epochs}}
\IF{subsample}
    \STATE $\text{sampledist = randomizedsample()}$
\ENDIF
\IF{shuffle}
    \STATE $\text{permute sampledist}$
\ENDIF
\FOR{$j=1$ to $\text{len(sampledist})$}
\STATE A = sampledist[j]
\STATE dat = rand(k,n,1)
\STATE output = batchmultiply(A, dat)
\STATE prev = mask(model)
\STATE train(model, dat, output, k, batchsize, epochs)
\STATE model = merge(prev)
\ENDFOR
\ENDFOR
\end{algorithmic}
\caption{Train}
\label{alg:seq}
\end{algorithm}
\noindent Binary symmetric matrices compose the training data  $d$ and it is distributed proportionally to its sample space with respect to the matrix elementwise sum or L1-norm. To generate $d$ we compute the likelihood $p(x)$ of observing a matrix with a given L1-norm x, then randomly sample $p(x)*|d|$ matrices with said sum for all possible norms. Training variation is reduced with symmetry because $|d| = 36n^{1.3}$ is a rapidly decaying fraction of the space of mask matrices, which without truncating the sample space $p(x)$ at high $n$ would be spread too thin.
\section{Results}
\begin{figure}[hbp]
\captionsetup[subfloat]{farskip=0pt,captionskip=-7.8pt}
    \captionsetup[subfigure]{labelformat=empty}
    \centering
  \subfloat[]
  {\includegraphics[width=0.43\textwidth]{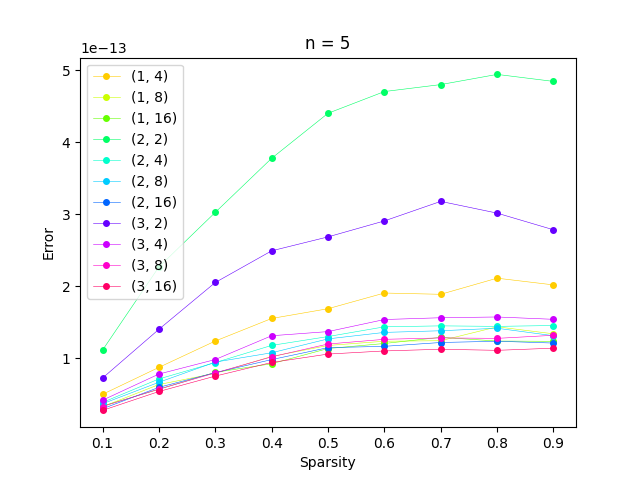}\label{fig:f1}}
  \hfill
  \subfloat[]
  {\includegraphics[width=0.43\textwidth]{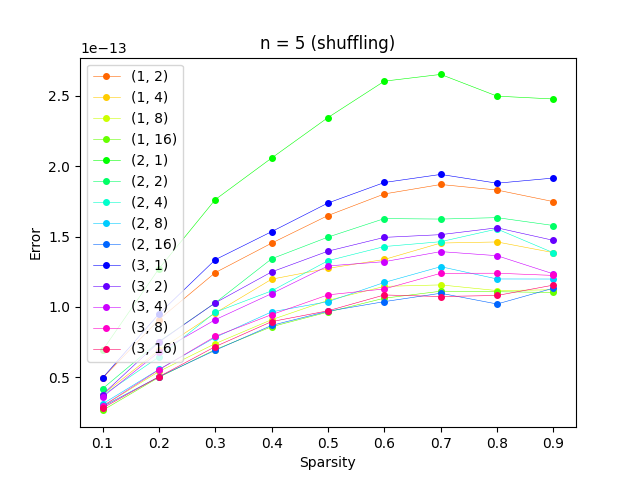}\label{fig:f2}}\\[-2ex]

  \subfloat[]
  {\includegraphics[width=0.43\textwidth]{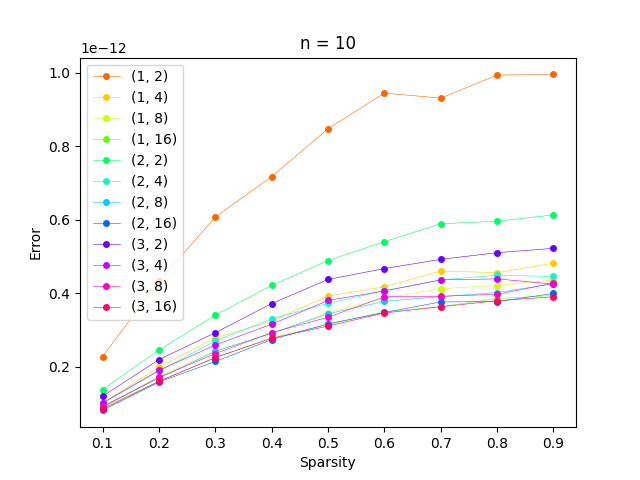}\label{fig:f1}}
  \hfill
  \subfloat[]
  {\includegraphics[width=0.43\textwidth]{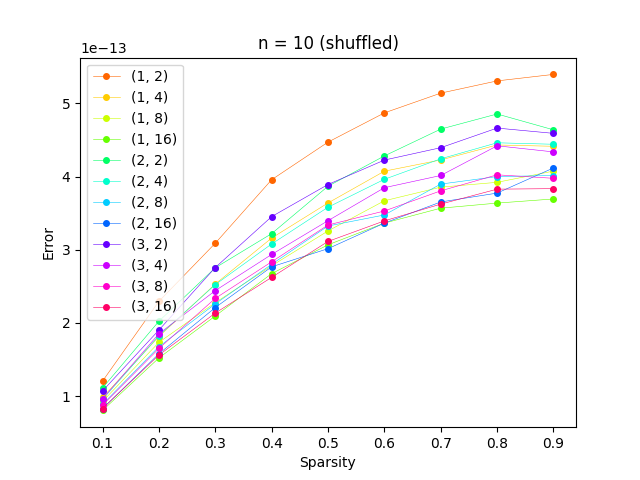}\label{fig:f2}}\\[-2ex]
 
  \subfloat[]
  {\includegraphics[width=0.43\textwidth]{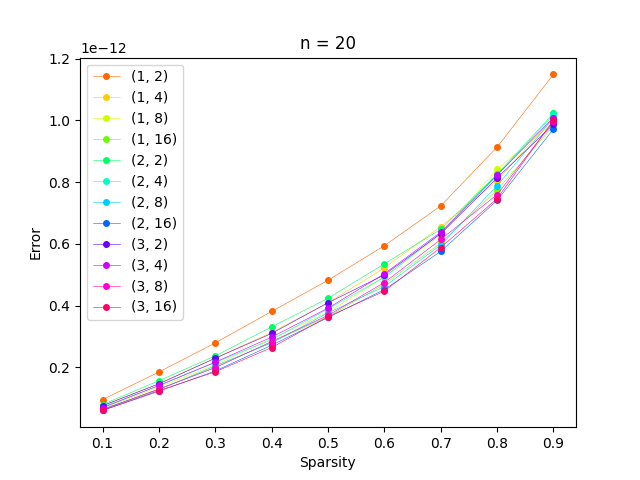}\label{fig:f1}}
  \hfill
  \subfloat[]
  {\includegraphics[width=0.43\textwidth]{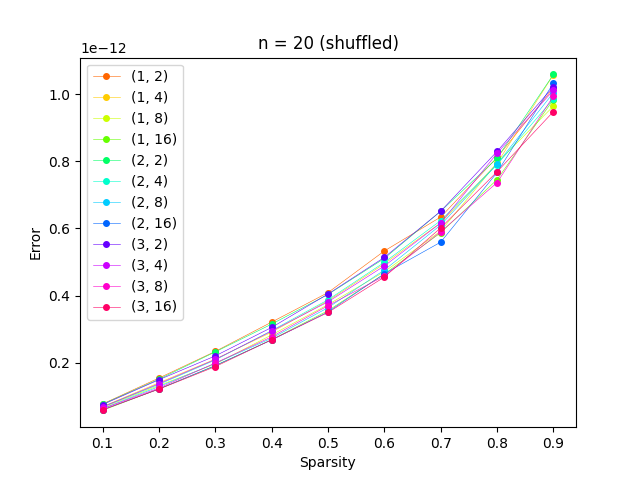}\label{fig:f2}}\\[-2ex]

\subfloat[]
  {\includegraphics[width=0.43\textwidth]{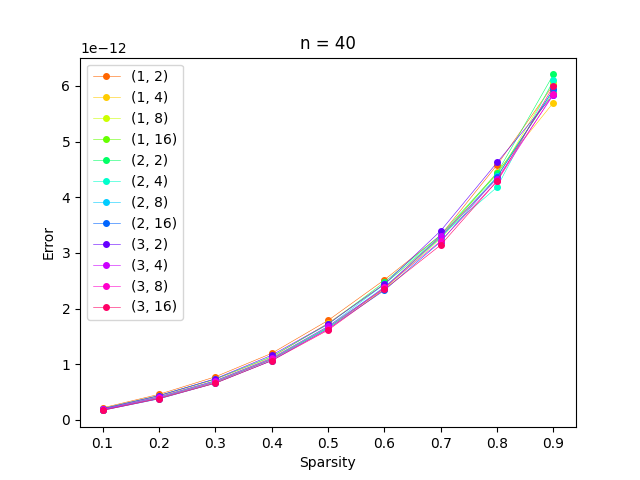}\label{fig:f1}}
  \hfill
  \subfloat[]
  {\includegraphics[width=0.43\textwidth]{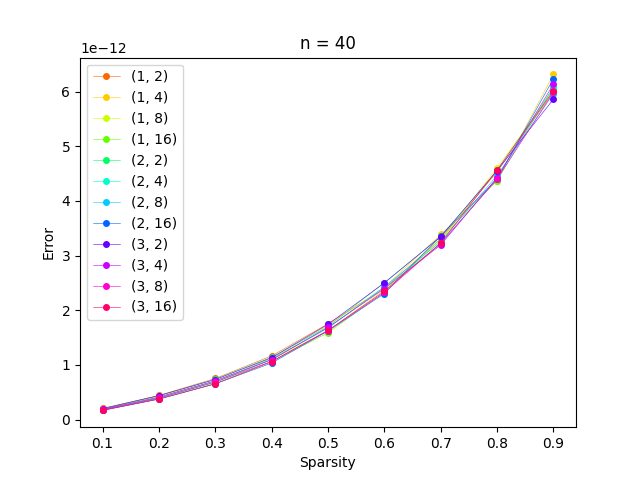}\label{fig:f2}}\\[-2ex]

    \subfloat[]
  {\includegraphics[width=0.43\textwidth]{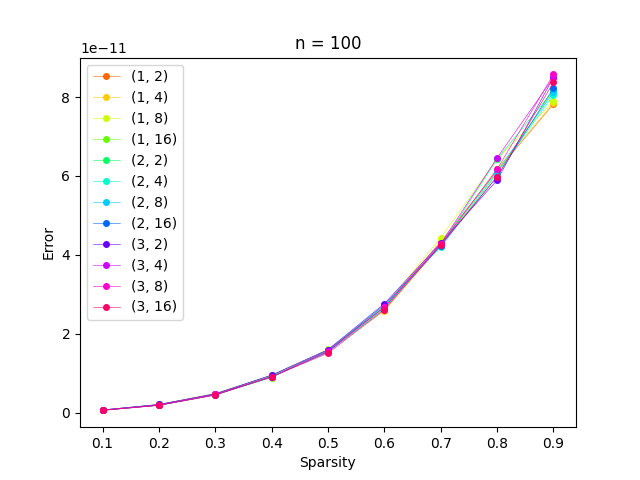}\label{fig:f1}}
  \hfill
  \subfloat[]
  {\includegraphics[width=0.43\textwidth]{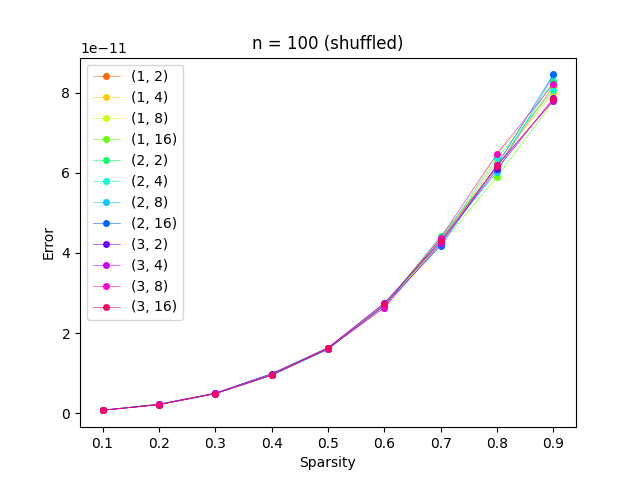}\label{fig:f2}}\\[-2ex]
  \caption{Average error by sparsity (average fraction of non-zero entries) for n=5,10,20,40,100. n varies by row, and experiment type by column. }
  \label{fig:sparse}
\end{figure}

Table 1 demonstrates network convergence on continuous $A$ as errors are near 0 for control and shuffled conditions. The binary errors reported in Fig. \ref{fig:sparse} at 0.5 sparsity are roughly equivalent to those of Table 1, so we can conclude the network pairs matrix indices with outgoing edges from the input layer since differentially weighting entries $(i,j), (j,i)$ does not significantly impact error. Convergence occurs quickly in this high training data regime as every (epoch, meta-epoch) pair converged for n=20 and higher, while for n=10 all (x,1) pairs for $x \in [1,2,3]$ failed to converge, as did n=5 but including (1,2). 2 meta-epochs and above generally got the control condition working, while for the shuffled condition only (1,1) for n=5 failed. An inspection of the learned weights (Fig. \ref{fig:weights}) details how the network converges on the fixed operator and explains why there is no significant variation between binary and continuous data. With masking, this problem is trivial and amounts to learning the all-ones-matrix since we grant ourselves access to scaling the weight matrices directly.
\begin{table}[h]
	\centering
	\caption{Minimum parameter sweep errors and corresponding (epoch, meta-epoch) values over unsymmetric test data with $A_{ij} \in [0,1]$ at 0.5 sparsity.}
	\label{table1}
	\begin{tabular}{@{}cccc@{}}
		\textbf{n, method}& \textbf{Control}& \textbf{Shuffled}& \textbf{Randomly Subsampled} \\
		 \textbf{5} & 6.36e-14 (1,16) & \cellcolor{green!20}5.32e-14 (3,16) & 0.045 (1,8)\\
		\textbf{10} & 1.56e-13 (1,16) & \cellcolor{green!20}1.51e-13 (3,16) & 0.052 (2,8) \\
		\textbf{20} & 4.77e-13 (3,16) & \cellcolor{green!20}1.95e-13 (3,8) & 0.046 (2,16) \\
		\textbf{40} & 2.065e-12 (2,8) & \cellcolor{green!20}1.98e-12 (1,8) & 0.055 (2,4) \\
		\textbf{100} & \cellcolor{green!20}1.51e-11 (3,16) & 1.61e-11 (3,16) & 0.16 (2,1) \\
	\end{tabular}
\end{table}
\newline
Previous experiments with $|d| = 36n^{1.15}$ and $k=15n^{\frac{2}{3}}$ admitted much more spread across the parameter sweep and random sampling performed well if not better than the other two conditions. However, at high n, the likelihood $p(x)$ of observing a matrix with a given L1-norm was so negligible that test errors could not easily be computed with the programs written and so an analysis at lower data regimes was omitted in this study. Fig. \ref{fig:sparse} omits random subsampling due to poor performance, but otherwise measures binary matrix error as a function of sparsity and we can see that the spread decreases with high n and a quadratic or exponential error scaling emerges likely due to having $n^2$ entries that the learned weights can slightly vary over. 
\newpage
\begin{figure}[hbp]
\[
() \begin{bsmallmatrix}
1 & 1 & \hdots & 1 \\
1 & 1 & \hdots & 1 \\
\vdots & \vdots & \ddots & \vdots\\ 
1 & 1 & \hdots & 1
\end{bsmallmatrix}
\begin{bsmallmatrix}
x_1 \\ x_2 \\ \vdots \\ x_n 
\end{bsmallmatrix} = \begin{bsmallmatrix} \sum_{1}^{n}x_{i}\\ \sum_{1}^{n}x_{i} \\ \vdots \\ \sum_{1}^{n}x_{i} \end{bsmallmatrix}
\]\[
(1) \begin{bsmallmatrix}
\frac{1}{c_1} & & & \\ & \frac{1}{c_2} & & & \\ & & \ddots & \\ & & & \frac{1}{c_n}
\end{bsmallmatrix}
\begin{bsmallmatrix}
c_1 & c_2 & \hdots & c_1 \\
c_2 & c_2 & \hdots & c_2 \\
\vdots & \vdots & \ddots & \vdots\\ 
c_n & c_n & \hdots & c_n
\end{bsmallmatrix}
\begin{bsmallmatrix}
x_1 \\ x_2 \\ \vdots \\ x_n 
\end{bsmallmatrix} = \begin{bsmallmatrix} \sum_{1}^{n}x_{i}\\ \sum_{1}^{n}x_{i} \\ \vdots \\ \sum_{1}^{n}x_{i} \end{bsmallmatrix}
\]\[(1,1) \begin{bsmallmatrix}
\sqrt{\frac{1}{c_1}} & & & \\ & \sqrt{\frac{1}{c_2}} & & & \\ & & \ddots & \\ & & & \sqrt{\frac{1}{c_n}}
\end{bsmallmatrix}^{2}
\begin{bsmallmatrix}
c_1 & c_2 & \hdots & c_1 \\
c_2 & c_2 & \hdots & c_2 \\
\vdots & \vdots & \ddots & \vdots\\ 
c_n & c_n & \hdots & c_n
\end{bsmallmatrix}
\begin{bsmallmatrix}
x_1 \\ x_2 \\ \vdots \\ x_n 
\end{bsmallmatrix} = \begin{bsmallmatrix} \sum_{1}^{n}x_{i}\\ \sum_{1}^{n}x_{i} \\ \vdots \\ \sum_{1}^{n}x_{i} \end{bsmallmatrix}\]

\caption{Empirically learned weights for network. Constants can be positive or negative and decrease in magnitude with increasing n. Architectures with multiple copies such as $(2)$ learn less discernable weights but perform well.}
\label{fig:weights}
\end{figure}
\noindent Low training data regimes benefit from increased exposure to the sample space with subsampling, and gain more from increasing epochs due to being further from convergence at low parameter pairs. Both benefit from sampling, and high regimes encounter less drawback from increasing meta-epochs whereas the former can get over-exposed to a sufficiently small sample and fail to generalize. High regimes get overwhelmed with repeated large subsamples. $(1)$ of figure \ref{fig:weights} clearly indicates how subpaths of the network are learning and separating each row-wise multiplication composing the target function as seen by the differing constants. A $()$ architecture with no hidden layers takes less computation to train, but $(1)$ may imply less gradient updates are required to converge from randomly initialized weights due to having an infinite number of settling points whereas $0$ has one. Increasing width and depth decreases transparency of the learned weights but may be needed for approximating a more complex fixed operation than matrix-vector multiplication. 

\section{Conclusion}
The two mechanisms introduced - semantically tying masks to edge weights and hence edge weights to the algebra of the target fixed operation, and network pruning to respect the mask's dependency structure while still leaving space for deciding network width and depth - find utility on an overlooked base case of matrix-vector multiplication $\phi(A,x) \rightarrow Ax$. Given varied $A,x$ input, $Ax$ outputs, and one masking per input, a randomly initialized network learned the target fixed operation. Furthermore it does so with any hidden layer count and size mod n without scaling with anything other than information from $A$.
\newline\\
Masking during backpropagation affords control over training and offers more degrees of freedom than there appear to be in a feedforward network. As long as the researcher understands how each layer is coupled - $h_{i} = \sigma_h(W_{i}(h_{i-1}))$ - one can modify every layer $W_{i}$ to distort the chain of multiplications (convolutions, etc.) with masking to achieve specific outcomes. While layer depth increases conditional dependence to previous layers and we get diminishing expressivity after the first layer, the fact that there are $\sim \prod_{i=1}^{m-1}2^{n_{i}}$ subnetworks with depth $m$ and edge counts $n_{i}$ encourages researchers not just to sparsify networks but also design training patterns to specific ends and imbibe meaning in the trained network's ensemble of subnetworks.

\section{Future Directions}
\begin{figure}[H]
{\includegraphics[width=.67\textwidth]{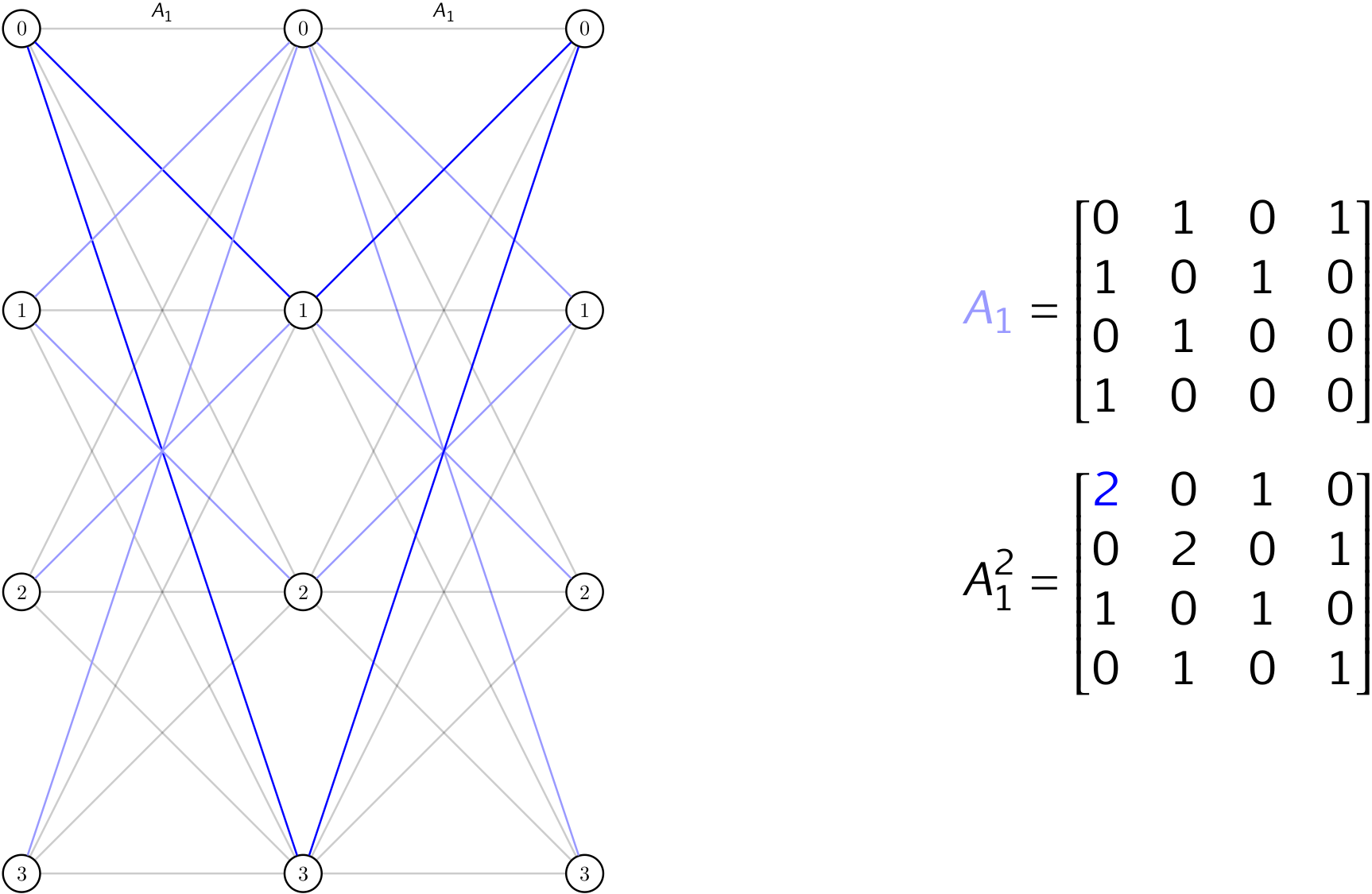}}
\caption{Suggested mechanic for indirect 2\textsuperscript{nd} order effects. By duplicating $A$ sequentially, each path multiplicity is correct accounting for the number of length-2 paths from $i$ to $j$. To capture both 1\textsuperscript{st} and 2\textsuperscript{nd} order effects, turn self-loops on in the second layer. Time varying $A$ can use two distinct masks.}
\label{fig:snd}
\end{figure}

Errors ranged with 3 orders of magnitude at $n=5$ for $(1,1)$ epoch, meta-epochs over different training samples indicating that specific training examples and input order have a strong effect on convergence time, encouraging a study quantifying the generalization/memorization quotient \cite{GM} of matrices and ordering effect to better refine training towards a fixed operation. There exists an input sequence which in the expectation yields best outcomes, and it is unclear how to find it but given the simplicity of the target function and low network dimension it may be a good place to look. To move from first-order interactions, multiple masking layers can be implemented (Figure \ref{fig:snd}). We suggest separating layers to account for the multiplicity of paths that reach a given node and to allow for lower-order effects to be taken into account.\newline\\
\begin{figure}[h]
{\includegraphics[width=.75\textwidth]{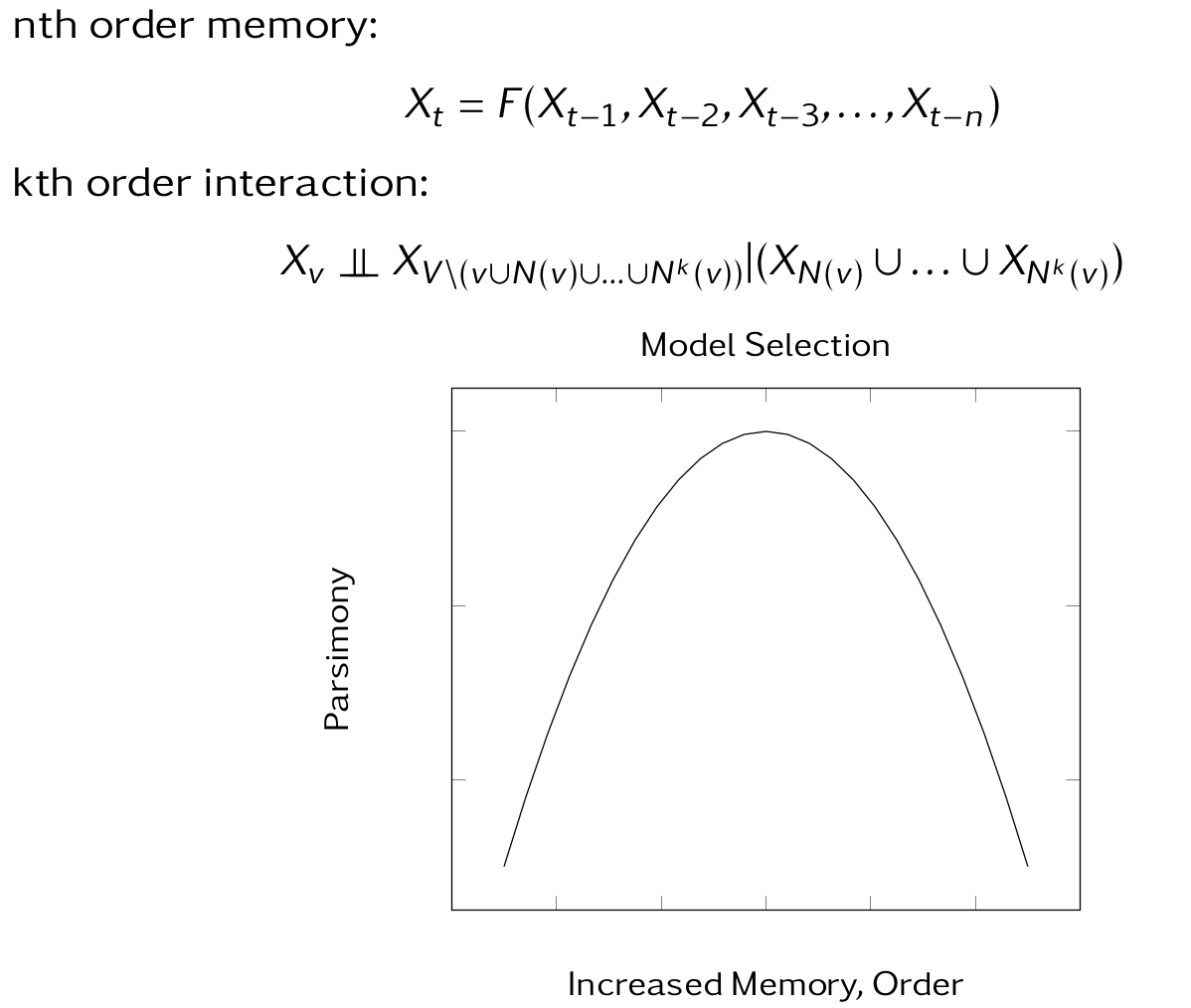}}
\caption{A general view of how increasing the amount of memory and neighbor interactions increases expressivity, and trying to fit data with the lowest order interaction and least memory could help to capture and understand the core dynamics of a system well.}
\label{fig:pars}
\end{figure}
\noindent To best act as a litmus-test for whether input data can be explained by a finite-order function, all additional machinery which preserves dependencies at the outputs should be included rather than just linear layers without activation functions to truly span the targeted subspace. Systems with a graphical basis often not only have interactions between components that have many degrees of separation whose interactions we hope to capture but also some memory (Figure \ref{fig:pars}). Recurrent networks such as LSTMs could assess whether memory is required to capture data well in a given domain. Broadly speaking, patterned masking could be used as a tool for escaping local minima, overcoming stickiness, and more generally directing traversal over the loss landscape - the masking sequence could be a target for another network to learn for example. Restricting a model's capacity for sufficiency testing is not limited to neural networks.\cite{foo}

\section{Acknowledgements}
This work utilized computing resources from the UCSB Center for Scientific Computing and Pacific Research Platform Nautilus cluster. The CSC staff in particular were helpful for troubleshooting issues. 

\bibliography{Ensemble_Mask_Networks}

\end{document}